\documentclass[10pt,twocolumn,letterpaper]{article}

\usepackage[pagenumbers]{cvpr} 

\usepackage{graphicx}
\usepackage{amsmath}
\usepackage{amssymb}
\usepackage{booktabs}
\usepackage{multirow}
\usepackage{booktabs}
\usepackage{stfloats}
\usepackage{url}
\usepackage{subcaption}
\usepackage[accsupp]{axessibility}  

\usepackage[pagebackref,breaklinks,colorlinks]{hyperref}

\usepackage[capitalize]{cleveref}
\crefname{section}{Sec.}{Secs.}
\Crefname{section}{Section}{Sections}
\Crefname{table}{Table}{Tables}
\crefname{table}{Tab.}{Tabs.}

\def\cvprPaperID{2845} 
\def\confName{CVPR}
\def\confYear{2022}

\begin{document}

\title{Global Sensing and Measurements Reuse for Image Compressed Sensing}

\author{Zi-En Fan, Feng Lian\thanks{Corresponding author.}, Jia-Ni Quan\\
Xi’an Jiaotong University, Xi’an, China\\
{\tt\small fze0012@stu.xjtu.edu.cn,lianfeng1981@mail.xjtu.edu.cn,jiani\_quan@stu.xjtu.edu.cn}
}
\maketitle

\begin{abstract}
	Recently, deep network-based image compressed sensing methods achieved high reconstruction quality and reduced computational overhead compared with traditional methods. However, existing methods obtain measurements only from partial features in the network and use them only once for image reconstruction. They ignore there are low, mid, and high-level features in the network\cite{zeiler2014visualizing} and all of them are essential for high-quality reconstruction. Moreover, using measurements only once may not be enough for extracting richer information from measurements. To address these issues, we propose a novel Measurements Reuse Convolutional Compressed Sensing Network (MR-CCSNet) which employs Global Sensing Module (GSM) to collect all level features for achieving an efficient sensing and Measurements Reuse Block (MRB) to reuse measurements multiple times on multi-scale. Finally, experimental results on three benchmark datasets show that our model can significantly outperform state-of-the-art methods.  Code is available at: \url{https://github.com/fze0012/MR-CCSNet}.

\end{abstract}

\section{Introduction}
\label{sec:intro}

Compressed Sensing \cite{donoho2006compressed}(CS), a signal processing technique, can efficiently acquire and reconstruct a signal. It has developed rapidly and attracted the attention of many researchers. Given a high-dimensional signal $x \in \mathbb{R}^{N}$, we get the measurements of $x$, denoted by $y\in \mathbb{R}^{M}$, by a linear mapping $y=\Phi x$, where $\Phi \in \mathbb{R}^{M \times N} $ is sensing matrix and $\frac{M}{N}$ is sampling ratio. Because $M \ll N$, recovering $x$ is generally impossible for the ill-posed problem. CS shows that the original signal $x$ can be recovered from $y$ with high probability when the signal $x$ is sparse in some domain\cite{candes2006near,duarte2008single}.

The two core problems of image CS are (1) the design of sensing matrix and (2) recovering the high-dimensional signal $x$ from its low-demensional measurements $y$. For the former one, many matrices \cite{gan2007block, amini2011deterministic, lu2017binary, dinh2013measurement,gao2015block} are proposed, but they are hand-designed and ignore there are statistical correlation between different elements of signal. For the latter one, the papers of \cite{li2013tval3, zhang2012image, zhang2014group, mun2009block, chen2011compressed} propose methods for exploring image priors and combining optimization criteria and iterative thresholding algorithms \cite{haupt2006signal}. These methods require high computational overhead and perform poorly when the sampling ratio is extremely low.

In recent years, deep learning has been widely used in computer vision and shows superior performance \cite{krizhevsky2012imagenet, long2015fully}. Researchers were inspired to solve these two challenges of CS with deep learning, called Deep Compressed Sensing (DCS). A few DCS methods \cite{shi2019image, sun2020dual, mousavi2018data, mousavi2017learning, zhang2018ista} have been proposed and achieve promising results since the powerful learning and representation capabilities of neural networks. 

Despite their success, existing DCS methods only use a convolutional layer to learn the sensing matrix, which ignores the spatial features in the image. In addition, because the residual architecture is widely used in reconstruction network, the reconstruction quality relies on it. To address these issues, Zheng \etal proposed RK-CCSNet \cite{zheng2020sequential}. For the former one, RK-CCSNet use the Sequential Convolutional Module (SCM) to gradually 
compact the image size through a sequence of filters. For the latter one, RK-CCSNet proposed the second-order residual architecture according to the relationship between ResNet \cite{he2016deep} and Ordinary Differential Equation.

Although RK-CCSNet proposed an effective strategy for image CS, it always suffers from these problems: (1) There are hierarchical nature of the features in the convolutional neural networks (CNNs): the low, mid, and high layer learn features such as edges, complex textures, and objects, respectively. But RK-CCSNet only samples from the highest layer, which ignores a large amount of rich features contained in those neglected layers; (2) Existing methods \cite{mousavi2017learning,zheng2020sequential,shi2019image, zhang2018ista} recover the original image from measurements using deep learning, which takes measurements as input and use it only once. It extracts features from measurements in a rather shallow manner.
\begin{figure}[t]
	\centering
	\includegraphics[width=1\linewidth]{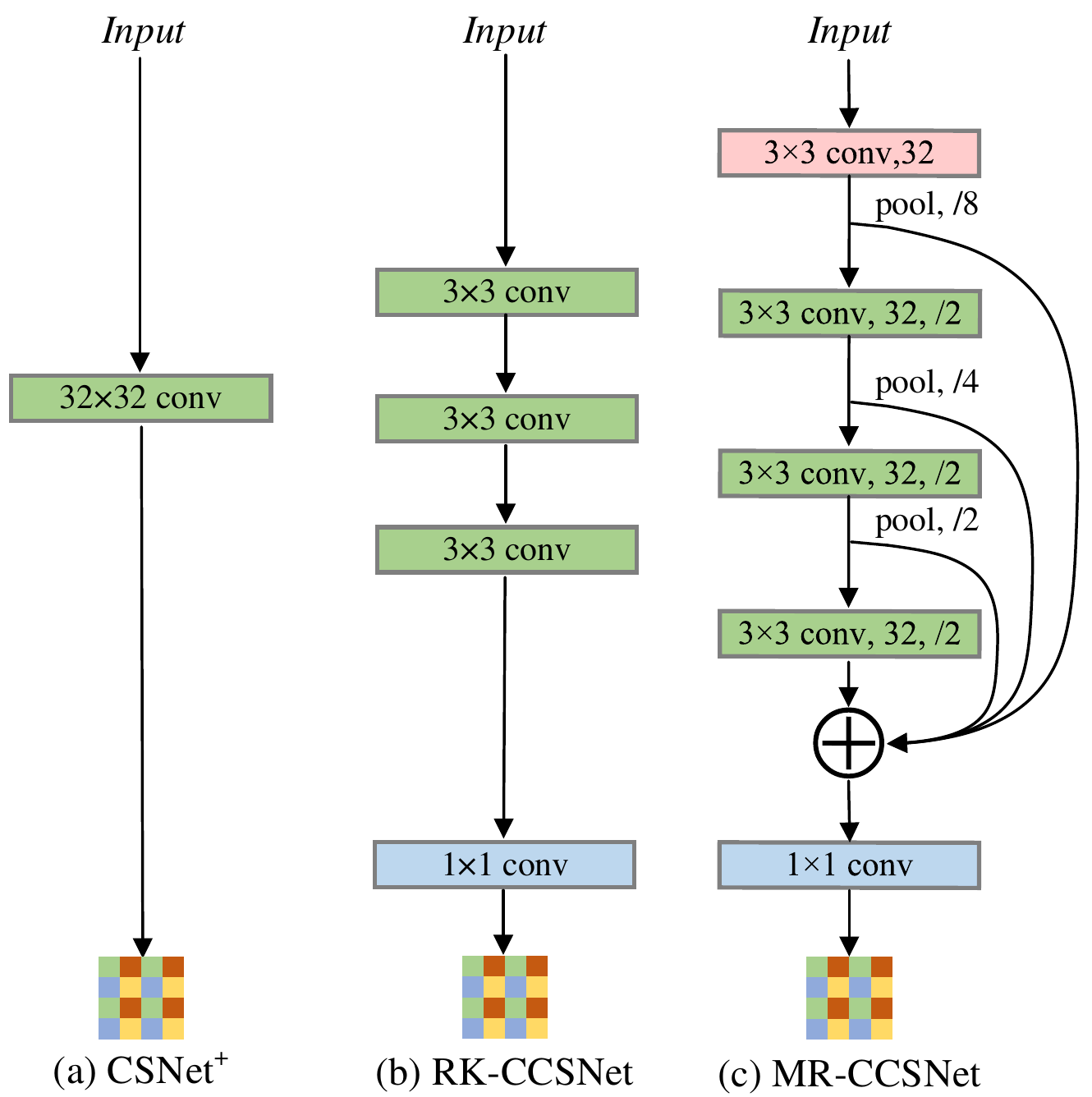}
	\caption{This figure shows the sensing network of CSNet$^{+}$ \cite{shi2019image}, RK-CCSNet \cite{zheng2020sequential}, and our proposed model. \textbf{(a)}: There is only a convolutional layer, so the spatial features in the image are ignored. \textbf{(b)}: Although there are multiple convolutional layers, it only samples from the highest layer. Therefore, there are the highest level features in the measurements and low and mid-level features are ignored. \textbf{(c)} Overview of GSM. To take advantage of the hierarchical nature of the CNNs, which are neglected in RK-CCSNet, we use a shortcut connection pass the features of different layer to the end. And then, $1 \times 1$ convolutions are used to sample. Because we sample from all level features, there are richer features in measurements than CSNet$^{+}$ and RK-CCSNet.}
	\label{shiyi}
\end{figure}

To address these issues, we propose Global Sensing Module (GSM) and Measurements Reuse Block (MRB). In GSM, as shown in \cref{shiyi}, we first use a convolutional layer to get a high dimensional feature. After that, we gradually compact the feature maps by multiple convolutional layers. And then we collect all level features in the network. Finally, we obtain measurements by $1 \times 1$ convolutions.
In order to match dimensions, we add pooling layer into the shortcut connection. 
In MRB, we first compact the phased reconstructed result and get multiple feature maps. After that, we extract matching information from measurements. Finally, we fuse them on multi-scale. It is a promising manner to move from shallow measurements utilizing to deep. 

We conduct experiments on three benchmark data sets: BSDS500 \cite{arbelaez2010contour}, Set5 \cite{bevilacqua2012low}, and Set14 \cite{zeyde2010single} and chose PSNR and SSIM \cite{hore2010image} as the evaluation metrics. Evaluation results indicate that MR-CCSNet can significantly outperform state-of-the-art methods. In particular, we show that our model can achieve high reconstruction quality at low sampling ratio. In addition, we show that GSM and MRB are effective by ablation studies.

To conclude, our contributions are three-fold: (1) proposal of the GSM which can achieve efficient sampling; (2) proposal of the MRB for making full use of measurements; (3) building an end-to-end network MR-CCSNet for image CS based on GSM and MRB, and demonstrating its effectiveness on three benchmark data sets.

\section{Related work}
\label{sec:rel}
CS aims to recover the original signal $x$ from its limited measurements $y$. We will review some of existing works from two categories: 
~\\

\noindent\textbf{Traditional Compressed Sensing}\quad Traditional CS methods regard the signal reconstruction as an optimization problem. The objective function of this optimization problem is
\begin{equation} 
	\label{eq:cs} 
	\centering
	\underset{x}{min}\frac{1}{2} \Vert\Phi x-y\Vert_{2} ^{2} + \lambda \Vert D x \Vert_{1},
\end{equation}
where $D$ is a projection operators (wavelet or DCT). $\ell_{1}$ norm controls the sparsity of $Dx$.

The convex optimization methods \cite{chen2001atomic}, the greedy algorithms \cite{mallat1993matching,tropp2007signal}, and the gradient descent methods \cite{daubechies2004iterative,wright2009sparse,figueiredo2007gradient} are three representative methods. For image compressed sensing, many researchers introduce other prior as a regularization item. In \cite{li2013tval3}, in order to improve the smoothness, Li \etal used the total variation (TV) regularized constraint. In \cite{zhang2014group}, Zhang \etal proposed group sparse representation (GSR), which sparsely represented images in
the domain of group and enforced the local sparsity
and nonlocal self-similarity. Furthermore, the projected Landweber (PL) algorithm\cite{bertero2020introduction} was introduced to image CS. In \cite{gan2007block}, Gan proposed block-based CS based Wiener filtering PL iteration.
In recent years, researchers have also proposed many improved PL-based methods \cite{chen2011compressed,mun2011residual,fowler2011multiscale}. Besides image reconstruction methods, some attention is also paid to the sensing matrix. In most works, the sensing matrix is a random Gaussian matrix, which satisfies the Restricted Isometry Property (RIP) \cite{candes2008restricted} with hith probability. Although so many methods have been proposed in traditional CS, they all require a lot of computing resources and perform poorly when the sampling ratio is extremely low.
~\\

\noindent\textbf{Deep Compressed Sensing}\quad The main idea of DCS is to learn the mapping from the measurements to the original signal using a neural network, so the speed and accuracy of reconstruction are improved. Generally, we train the network by minimizing the loss function
\begin{equation} 
	\label{eq:dcs} 
	\centering
	\underset{\theta}{min}\frac{1}{2} \sum_{n=1}^{N}\Vert x_{n} -F(y_{n},\theta)\Vert_{2}^{2},
\end{equation}
where the $x_{n}$ is the original image, $y_{n}$ is the measurements of $x_{n}$, and $F$ is the neural network parameterized by $\theta$.
Many DCS methods have been proposed \cite{mousavi2015deep, kulkarni2016reconnet,mousavi2017learning, shi2019scalable, zheng2020sequential, shi2019image}.
In \cite{mousavi2015deep}, Mousavi \etal proposed a stacked denoising autoencoder (SDA) to learns a structured representation from training data. However, SDA has the computational complexity because it used many full connection layers. In \cite{kulkarni2016reconnet}, Kulkarni \etal introduced ReconNet, which used CNN to reduce the number of model parameters by weight sharing. Mousavi and Baraniuk \cite{mousavi2017learning} argued that real world signals are not exactly sparse on a fixed basis and the recovery algorithms take a lot of time to converge. And they proposed DeepInverse which learns both a effective representation for the signals and an inverse map. 
Shi \etal \cite{shi2019image} argued that these methods ignore the characteristics of signal and proposed a end-to-end model CSNet$^{+}$ which uses convolutional neural network in sampling and reconstruction stage. However, when the sampling ratios change, these methods needed to train the model again, which is difficult to deploy for practical applications. Hence, Shi \etal \cite{shi2019scalable} attempted to solve this problem with greedy method and proposed SCSNet. Mousavi \etal \cite{mousavi2018data} proposed DeepSSRR which employs a parallelization scheme in the signal sensing and recovery process to accelerate the convergence speed. 
\begin{figure*}[t]
	\centering
	\includegraphics[width=.9\linewidth]{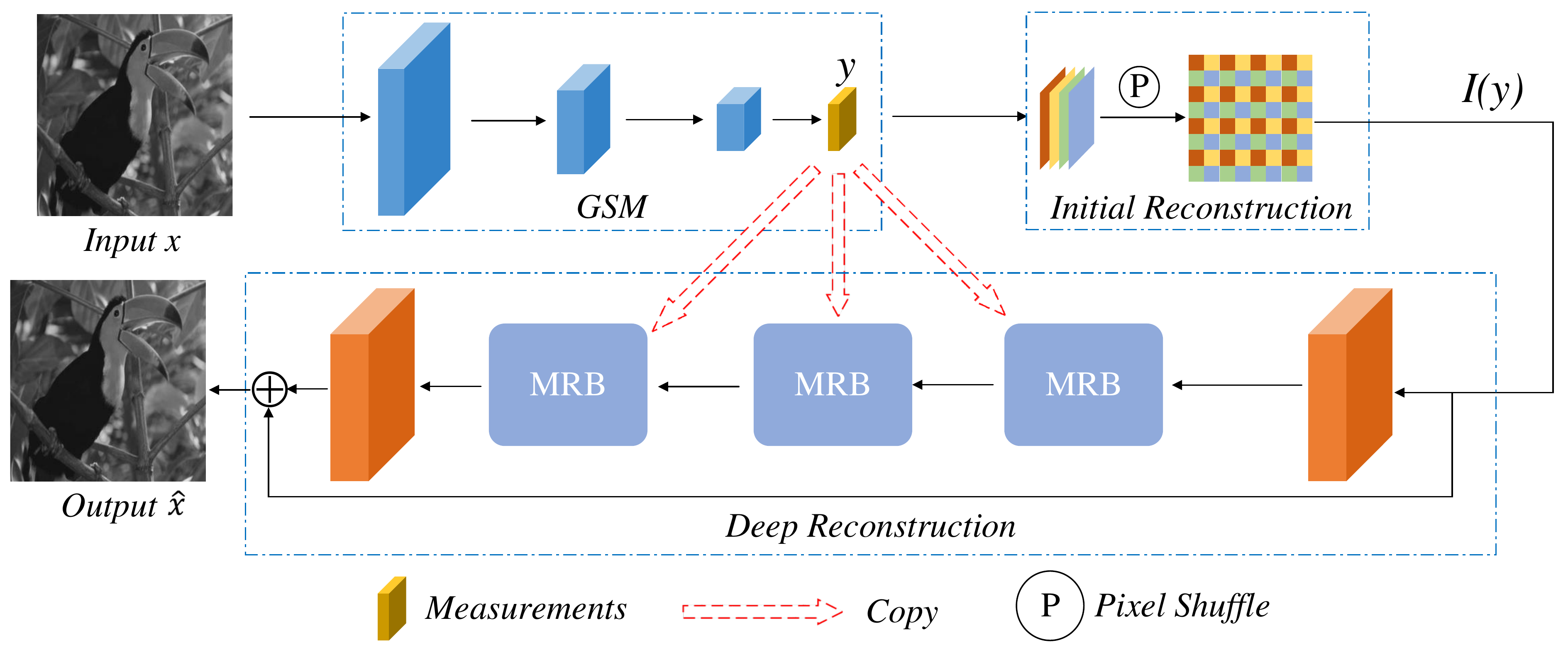}
	\caption{Overview of the proposed model. For the original image $x$, we obtain measurements $y$ from the sensing network GSM. And then the initial reconstructed image $I(y)$ is generated by the initial reconstruction network. Finally, we refine the $I(y)$ by the deep reconstruction network.}
	\label{H-CCSNet archi}
\end{figure*} 
In \cite{zheng2020sequential}, Zheng argued that existing end-to-end methods do not preserve the spatial features in the image and proposed RK-CCSNet, which applies Sequential Convolutional Module (SCM) to gradually compact measurements through a series of convolution filters. In addition, RK-CCSNet also proposed a novel Learned Runge-Kutta Block (LRKB) based on the famous Runge-Kutta methods for improving the reconstruction quality. 

Our work is also inspired by the idea of multi-scale in image processing. In \cite{xu2018lapran}, Xu \etal proposed a Laplacian pyramid reconstructive adversarial network (LAPRAN) which reconstructs the original image through multiple stages with different resolution simultaneously. Our model also employs MRB to fuse features learned from measurements on multi-scale.

\section{Methodology}
In this section, we will introduce our model in the case of sampling ratio is 6.25\%. \cref{H-CCSNet archi} shows the architecture of MR-CCSNet. Following CSNet$^{+}$ \cite{shi2019image} and RK-CCSNet \cite{zheng2020sequential}, MR-CCSNet has a sensing network GSM, an initial reconstruction network, and a deep reconstruction network. We denote them as $S(\cdot)$, $I(\cdot)$, $D(\cdot)$, respectively. Firstly, we obtain the measurements $y$ from $S(\cdot)$. Then, we use a linear mapping $I(\cdot)$ to generate initial reconstructed image $I(y)$. Because the quality of $I(y)$ is not enough, we refine it by a non-linear network $D(\cdot)$. To move from shallow measurements utilizing to deep, we stack multiple MRBs in the deep reconstruction network. 

In the sensing network $S(\cdot)$, we directly use convolutional layers for the whole images instead of dividing the images into non-overlapping block \cite{shi2019image,zheng2020sequential, shi2019scalable}. For satisfying the linear property, there is no bias and activation function in the network. This process can be written as:
\begin{equation} 
	\label{eq:first stage} 
	\centering
	y=S(x),
\end{equation}
where $x \in \mathbb{R}^{1 \times H \times W} $ and $y \in \mathbb{R}^{4 \times \frac{H}{8} \times \frac{W}{8}}$.

In $I(\cdot)$, the depth-wise convolution layer expands the measurements in channel dimension and the shape becomes $64 \times \frac{H}{8} \times \frac{W}{8}$.  
Then we get a $1 \times H \times W$ tensor by a pixel shuffle layer. This is the first time to utilize the measurements.

In the deep reconstruction network $D(\cdot)$, we first convert the initial reconstructed $I(y)$ image to a high dimensional feature by a convolutional layer. Then repeated MRBs, which share the same internal structure, are used to fuse them with matching features extracted from measurements $y$ multiple times on multi-scale. This is the second time to utilize the measurements. 

Finally, we use a convolutional layer to get the reconstructed image. In addition, we add a shortcut connection to the deep reconstruction network. The final reconstructed image $\hat{x}$ can be written as:
\begin{equation} 
	\label{eq:second stage} 
	\centering
	\hat{x}=D(I(y))+I(y)
\end{equation}

Our model uses two novel modules, GSM and MRB. They are explained below.
\subsection{Global Sensing Module}
\label{hscana}

\begin{figure}
	\centering
	\begin{subfigure}[b]{0.19\textwidth}
		\includegraphics[width=\textwidth]{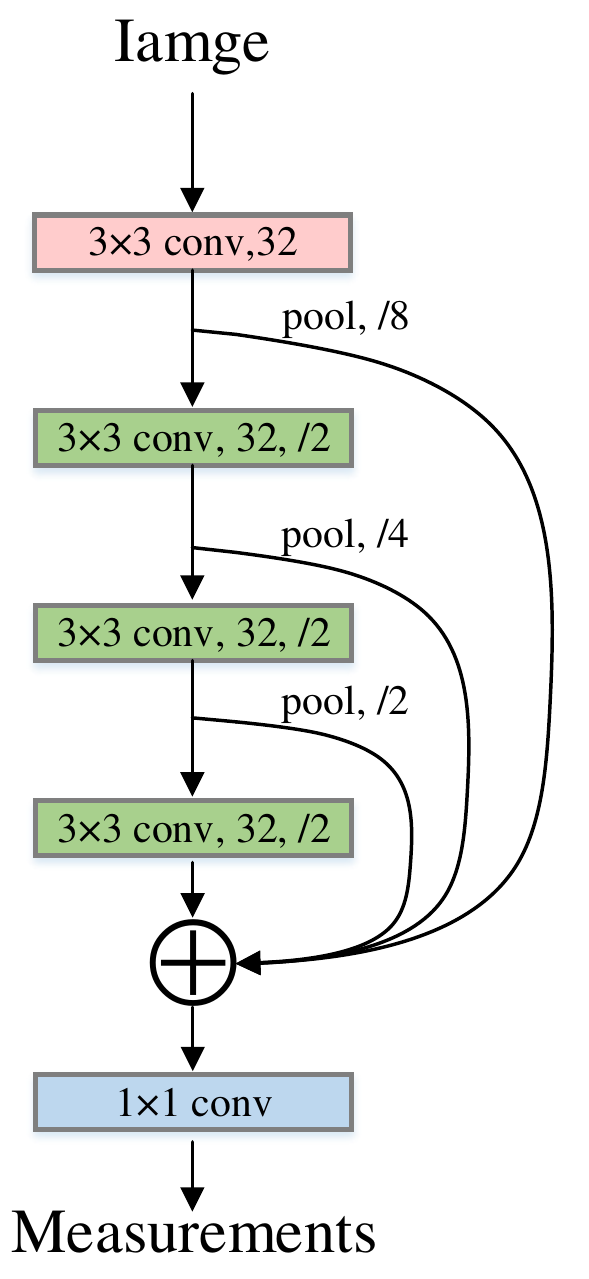}
		\caption{GSM\qquad \qquad \quad}
		\label{originalhsc}
	\end{subfigure}
	~
	\begin{subfigure}[b]{0.19\textwidth}
		\includegraphics[width=\textwidth]{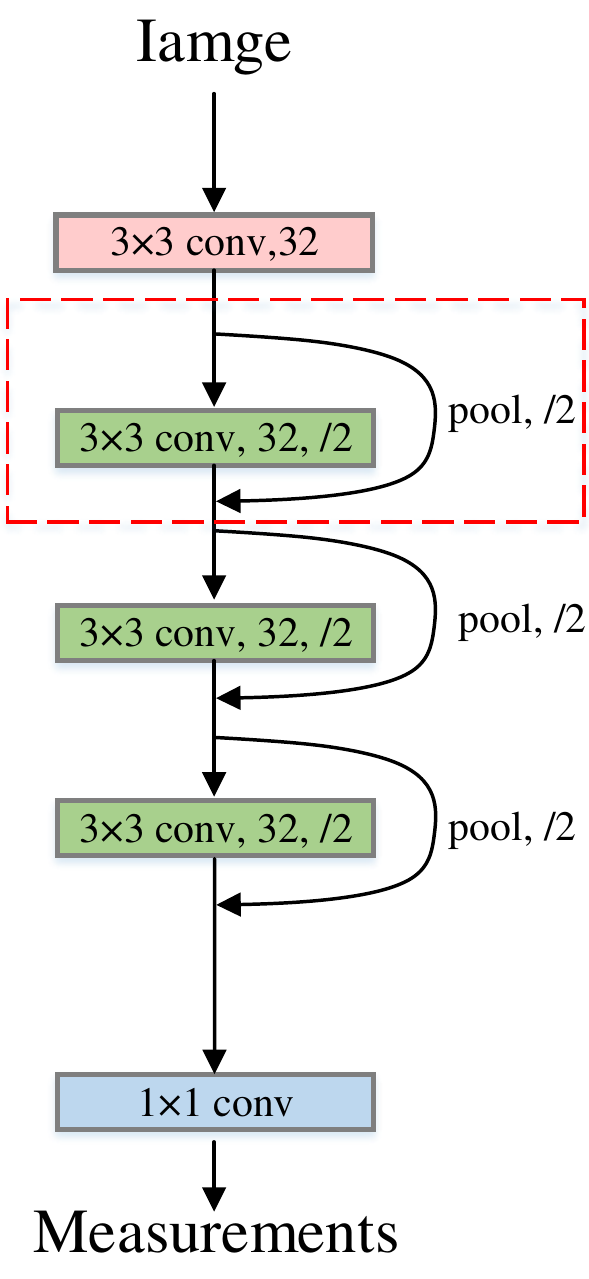}
		\caption{GSM$^{+}$\qquad \qquad}
		\label{simplifiedhsc}
	\end{subfigure}
	\caption{Comparison of the GSM and the GSM$^{+}$. The GSM$^{+}$ is flexible and can be easily used at various sampling ratios by repeating the building block, which is marked with red dotted box.}
\end{figure}

By analyzing existing methods, we argue that a good feature extraction network can help sample. In addition, we learn that convolutional neural networks extract features in a hierarchical manner which means layers close to the input to learn low-level features, like lines and simple textures, and layers deeper in the model to learn high-order features, like shapes or specific objects from \cite{zeiler2014visualizing}. Based on these two principles, our proposed method GSM, as shown in \cref{originalhsc}, has two stages. In the first stage, we use $3\times 3$ convolution layers to extract features. In the second stage, we collect all level features in the network and use a $1\times 1$ convolution layer to sample, rather than only from the low features (\ie CSNet$^{+}$) or high features (\ie RK-CCSNet).

In GSM, to collect all level features for sampling, we use a shortcut connection to pass the features of different layers to the end, and the pooling layer is added for matching the dimensions. 

When the sampling ratio changes, the GSM is not flexible for meeting the new requirements. Inspired by ResNet \cite{he2016deep}, we propose the GSM$^{+}$, as shown in \cref{simplifiedhsc}. Different from GSM, we add a shortcut connection between two successive layers rather than add it from different layers to the end directly. The building block of GSM$^{+}$ is marked with red dotted box and defined as:
\begin{equation} 
	\label{eq:sample} 
	\centering
	y_{t+1}=Conv(y_{t}) + P(y_{t}),	
\end{equation}
where $Conv$ and $P$ denote convolution layer and mean-pooling layer, respectively.
The sampling ratio is controlled by the number of building block and the blue block, so it is flexible and can be easily used at various sampling ratios by repeating the building block. 

In GSM$^{+}$, we can observe that it collect all level features for sampling, which is equivalent to GSM, by an iterative manner. Furthermore, there are richer features than GSM at each layer, because the features from former layer are passed to the current layer by shortcut connections. In a way, it achieve a more efficient feature extraction. When the sampling ratio is 50\%, there is only one building block in GSM$^{+}$, so GSM$^{+}$ degenerate into GSM. As the sampling ratio decreases, GSM is a special form of GSM$^{+}$.

In the CS theory, the measurements is obtained by a linear mapping. It is trivial that the convolution layer and the mean-pooling layer are linear mappings. So the building block is linear mapping. According to composition preserves linearity, the GSM$^{+}$ is a linear mapping.

\subsection{Measurements Reuse Block} 
The measurements are used only once for image reconstruction, which is difficult to extract richer information from measurements. The goal of MRB is to explore a novel approach for making full use of measurements multiple times on multi-scale. 

\begin{figure*}[t]
	\centering
	\includegraphics[width=.9\linewidth]{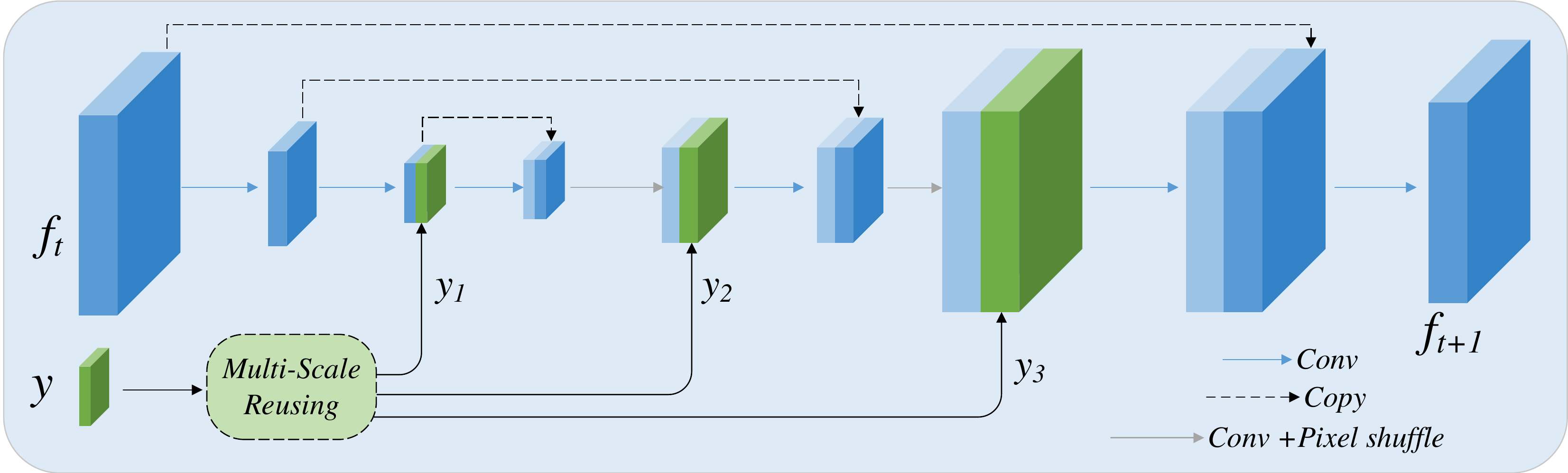}
	\caption{Overview of the proposed MRB.}
	\label{mrb-a}
\end{figure*}

\cref{mrb-a} illustrates the architecture of MRB. Phased reconstructed result $\mathnormal{f}_{t} \in \mathbb{R}^{C\times H\times W}$ and measurements $y \in \mathbb{R}^{C\times \frac{H}{4}\times \frac{W}{4}}$ are fed into MRB. 
\begin{figure}[t]
	\centering
	\includegraphics[width=0.9\linewidth]{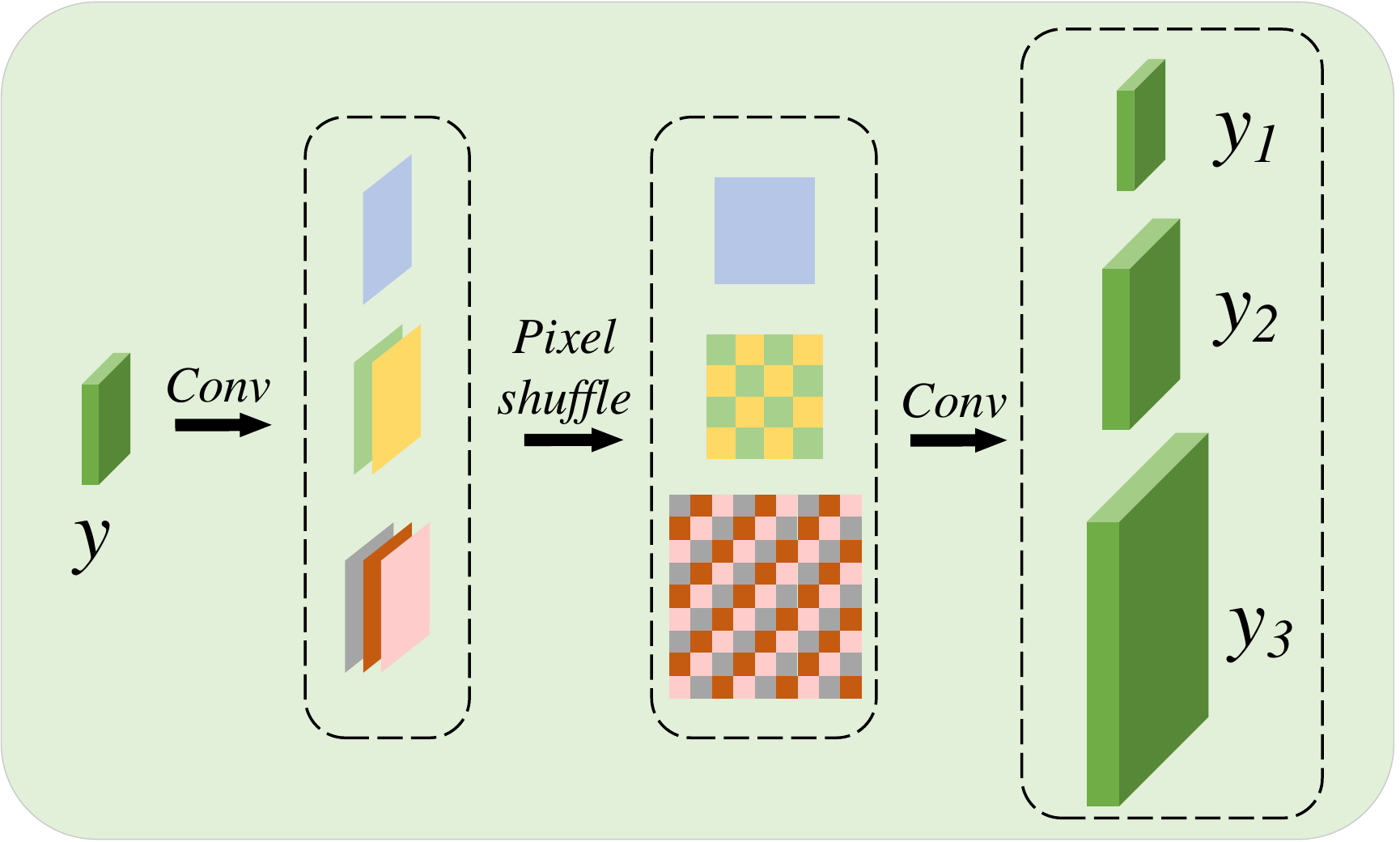}
	\caption{Overview of Multi-Scale Reusing. It aims to extract matching information for the backbone network of MRB.}
	\label{mrb-b}
\end{figure}
We firstly use two convolutional layers, denoted as $Conv_{1}$ and $Conv_{2}$, to obtain a compacted feature map $\mathnormal{f}^{\downarrow}$ and $\mathnormal{f}^{\downdownarrows}$. This process can be written as:
\begin{equation} 
	\label{eq:fuse1} 
	\centering
	f^{\downarrow}=Conv_{1} (f_{t}),	
\end{equation}
\begin{equation} 
	\label{eq:fuse2} 
	\centering
	f^{\downdownarrows}=Conv_{2} (f^{\downarrow}),	
\end{equation}
where $f^{\downarrow} \in \mathbb{R}^{C \times \frac{H}{2} \times \frac{W}{2}}$,  $f^{\downdownarrows} \in \mathbb{R}^{C \times \frac{H}{4} \times \frac{W}{4} }$. To fuse them with measurements on multi-scale, we then extract matching information from measurements and obtain three feature maps $y_{1}\in \mathbb{R}^{C \times \frac{H}{4} \times \frac{W}{4}}$, $y_{2}\in \mathbb{R}^{C \times \frac{H}{2} \times \frac{W}{2}}$, and $y_{3}\in \mathbb{R}^{C \times H \times W}$ by Multi-Scale Reusing, which is shown in \cref{mrb-b}. Next, $y_{1}$ is added into the backbone network of MRB and obtain $F_{1}$ by a concatenation operation and a convolutional layer. To preserve existing reconstruction results, we copy the $\mathnormal{f}^{\downdownarrows}$ again and fuse them with $F_{1}$ by a convolutional layer. Finally, a pixel shuffle layer is used to expand the fused feature map for next process.
This process can be written as:
\begin{equation} 
	\label{eq:fuse3} 
	\centering
	F_{1}=  Conv_{3}(f^{\downdownarrows}\oplus y_{1}),
\end{equation}
\begin{equation} 
 	\label{eq:fuse4} 
	\centering
	f^{\uparrow}=Pixel(Conv_{4}(F_{1}\oplus f^{\downdownarrows})),
\end{equation}
where $\oplus$ denotes a concatenation operation, $y\in \mathbb{R}^{2 \times \frac{H}{4} \times \frac{W}{4}}$, $y_{1}\in \mathbb{R}^{C\times \frac{H}{4} \times \frac{W}{4}}$, $F_{1} \in \mathbb{R}^{C \times \frac{H}{4} \times \frac{W}{4}}$ and $f^{\uparrow} \in \mathbb{R}^{C \times \frac{H}{2} \times \frac{W}{2}}$. By repeating this process, phased reconstructed result measurements and measurements are fused at multi-scale. Then the output $f_{t+1} \in \mathbb{R}^{C \times H \times W}$ is utilized as the input of next operation.

The MRB is not only a promising way for improving utilization of measurements, but also refines the phased reconstructed result at multi-scale. 

\subsection{Loss function}
In the training phase, we use the mean square error to measure the reconstruction quality. Specifically, for the initial reconstruction network, the loss function can be written as :
\begin{equation} 
	\label{loss initial} 
	\centering
	\mathnormal{l}_{int}=\sum_{k=1}^n \Vert I(S(y_{k};\theta);\phi_{int})-x_{k} \Vert_{F}^2.
\end{equation}
For the deep reconstruction network, the loss function can be written as :
\begin{equation} 
	\label{loss deep} 
	\centering
	\mathnormal{l}_{deep}=\sum_{k=1}^n \Vert D(I(S(y_{k};\theta);\phi_{int});\phi_{deep})-x_{k} \Vert_{F}^2,
\end{equation}
where the $\theta$, $\phi_{int}$, and $\phi_{deep}$ denote the parameters of the sensing network $S(\cdot)$, the initial reconstructed network $I(\cdot)$, and the deep reconstructed network $D(\cdot)$, respectively.
Therefore, we define a loss function of MR-CCSNet:
\begin{equation} 
	\label{loss} 
	\centering
	\mathnormal{l}=\mathnormal{l}_{deep}+\mathnormal{l}_{int}.
\end{equation}
\section{Experiments}
\subsection{Datasets and implementation details}
Following RK-CCSNet \cite{zheng2020sequential}, we use 400 images from BSDS500 \cite{arbelaez2010contour} dataset to train our model. For testing, we report the performance on three standard benchmark datasets: Set5 \cite{bevilacqua2012low}, Set14 \cite{zeyde2010single}, and BSDS100 \cite{arbelaez2010contour}\footnote{https://www2.eecs.berkeley.edu/Research/Projects/CS/vision/bsds/}. We convert these images into YCbCr space and the Y channel is used as the input for training and testing. During training, in order to increase the number of samples, we randomly crop the image with patch size 96$\times$96, and randomly flip horizontally. During testing, because the size of these images is inconsistent, we resize the image from Set5 and Set14 into 256$\times$256 and the image from BSDS100 into 480$\times$320. To optimize our model, we use Adam optimizer \cite{kingma2014adam} with $\beta_{1}=0.9$, $\beta_{2}=0.999$. The batch size is set to 4 and our model is trained for 200 epochs. The initial learning rate is set to $10^{-3}$ and reduced to quarter at 60, 90, 120, 150 and 180 epochs respectively. Six sampling ratios, \ie 1.5625\%, 3.1250\%, 6.2500\%, 12.5000\%, 25.0000\%, and 50.0000\% are investigated. PSNR (Peak Signal-to-Noise Ratio) and SSIM (Structural SIMilarity)\cite{hore2010image} are chosen as the evaluation metrics. We implement the model using PyTorch, and train it on Nvidia RTX 2080Ti GPU.
\subsection{Comparison with the state-of-the-arts}
To verify the effectiveness of MR-CCSNet and MR-CCSNet$^{+}$ where the sensing network is GSM and GSM$^{+}$ respectively, we quantitatively and visually compare them with 4 state-of-the-art methods with available codes, which is \textbf{TVAL3} \cite{li2013tval3}, \textbf{GSR} \cite{zhang2014group}, \textbf{CSNet$^{+\S}$} \cite{shi2019image}, and \textbf{RK-CCSNet$^{\S}$}\cite{zheng2020sequential}\footnote{\S \space means that we use the code provided in \cite{zheng2020sequential} to train this model.}.


\begin{table*}[t]\small
	\centering
	\setlength{\tabcolsep}{5pt}{
	\begin{tabular}{cccccccccccccc}
		\toprule
		&                                 & \multicolumn{2}{c}{TVAL3}           & \multicolumn{2}{c}{GSR}             & \multicolumn{2}{c}{CSNet$^{+\S}$}          & \multicolumn{2}{c}{RK-CCSNet$^{\S}$}       & \multicolumn{2}{c}{MR-CCSNet}       & \multicolumn{2}{c}{MR-CCSNet$^{+}$} \\ \cline{3-14} 
		Data                                        & Ratio                           & PSNR  & SSIM                        & PSNR  & SSIM                        & PSNR  & SSIM                        & PSNR  & SSIM                        & PSNR  & SSIM                        & PSNR          & SSIM           \\ \hline
		\multicolumn{1}{c|}{\multirow{6}{*}{Set5}}  & \multicolumn{1}{c|}{1.5625\%}  & 19.00 & \multicolumn{1}{c|}{0.4844} & 21.39 & \multicolumn{1}{c|}{0.5815} & 24.45 & \multicolumn{1}{c|}{0.6360} & 25.31 & \multicolumn{1}{c|}{0.7033} & 25.72 & \multicolumn{1}{c|}{0.7193} & \textbf{25.79}         & \textbf{0.7189}        \\
		\multicolumn{1}{c|}{}                       & \multicolumn{1}{c|}{3.125\%}  & 19.89 & \multicolumn{1}{c|}{0.5415} & 23.70 & \multicolumn{1}{c|}{0.6822} & 27.19 & \multicolumn{1}{c|}{0.7666} & 27.79 & \multicolumn{1}{c|}{0.8061} & 28.19 & \multicolumn{1}{c|}{0.8174} & \textbf{28.27}         & \textbf{0.8208}         \\
		\multicolumn{1}{c|}{}                       & \multicolumn{1}{c|}{6.25\%}  & 22.03 & \multicolumn{1}{c|}{0.6175} & 27.59 & \multicolumn{1}{c|}{0.8163} & 28.68 & \multicolumn{1}{c|}{0.8002} & 30.63 & \multicolumn{1}{c|}{0.8799} & 31.10 & \multicolumn{1}{c|}{0.8901} & \textbf{31.25}         & \textbf{0.8918}         \\
		\multicolumn{1}{c|}{}                       & \multicolumn{1}{c|}{12.5\%} & 23.75 & \multicolumn{1}{c|}{0.7365} & 31.61 & \multicolumn{1}{c|}{0.9016} & 33.55 & \multicolumn{1}{c|}{0.9243} & 34.27 & \multicolumn{1}{c|}{0.9393} & 35.03 & \multicolumn{1}{c|}{0.9464} & \textbf{35.16}         & \textbf{0.9471}         \\
		\multicolumn{1}{c|}{}                       & \multicolumn{1}{c|}{25\%} & 27.39 & \multicolumn{1}{c|}{0.8522} & 36.32 & \multicolumn{1}{c|}{0.9510} & 37.69 & \multicolumn{1}{c|}{0.9650} & 38.04 & \multicolumn{1}{c|}{0.9712} & 39.24 & \multicolumn{1}{c|}{0.9761} & \textbf{39.37}         & \textbf{0.9766}         \\
		\multicolumn{1}{c|}{}                       & \multicolumn{1}{c|}{50\%} & 33.11 & \multicolumn{1}{c|}{0.9430} & 42.18 & \multicolumn{1}{c|}{0.9908} & 42.54 & \multicolumn{1}{c|}{0.9852} & 43.90 & \multicolumn{1}{c|}{0.9901} & 45.07 & \multicolumn{1}{c|}{0.9919} & \textbf{45.11}         & \textbf{0.9920}         \\ \hline
		\multicolumn{1}{c|}{\multirow{6}{*}{Set14}} & \multicolumn{1}{c|}{1.5625\%}  & 16.79 & \multicolumn{1}{c|}{0.3993} & 18.93 & \multicolumn{1}{c|}{0.4399} & 22.78 & \multicolumn{1}{c|}{0.5369} & 23.36 & \multicolumn{1}{c|}{0.5917} & 23.61 & \multicolumn{1}{c|}{0.5993} & \textbf{23.69}         & \textbf{0.6034}         \\
		\multicolumn{1}{c|}{}                       & \multicolumn{1}{c|}{3.125\%}  & 18.40 & \multicolumn{1}{c|}{0.4514} & 20.26 & \multicolumn{1}{c|}{0.5184} & 24.96 & \multicolumn{1}{c|}{0.6602} & 25.26 & \multicolumn{1}{c|}{0.6914} & 25.56 & \multicolumn{1}{c|}{0.6997} & \textbf{25.63}         & \textbf{0.7029}         \\
		\multicolumn{1}{c|}{}                       & \multicolumn{1}{c|}{6.25\%}  & 19.65 & \multicolumn{1}{c|}{0.5287} & 23.59 & \multicolumn{1}{c|}{0.6526} & 26.33 & \multicolumn{1}{c|}{0.7178} & 27.24 & \multicolumn{1}{c|}{0.7836} & 27.91 & \multicolumn{1}{c|}{0.7986} & \textbf{28.00}         & \textbf{0.7996}         \\
		\multicolumn{1}{c|}{}                       & \multicolumn{1}{c|}{12.5\%} & 21.03 & \multicolumn{1}{c|}{0.6379} & 28.08 & \multicolumn{1}{c|}{0.7915} & 30.12 & \multicolumn{1}{c|}{0.8610} & 30.42 & \multicolumn{1}{c|}{0.8798} & 30.97 & \multicolumn{1}{c|}{0.8889} & \textbf{31.06}         & \textbf{0.8898}         \\
		\multicolumn{1}{c|}{}                       & \multicolumn{1}{c|}{25\%} & 22.69 & \multicolumn{1}{c|}{0.7731} & 31.82 & \multicolumn{1}{c|}{0.8939} & 33.81 & \multicolumn{1}{c|}{0.9339} & 34.16 & \multicolumn{1}{c|}{0.9443} & 35.04 & \multicolumn{1}{c|}{0.9510} & \textbf{35.11}         & \textbf{0.9512}         \\
		\multicolumn{1}{c|}{}                       & \multicolumn{1}{c|}{50\%} & 26.61 & \multicolumn{1}{c|}{0.9004} & 37.47 & \multicolumn{1}{c|}{0.9619} & 38.59 & \multicolumn{1}{c|}{0.9752} & 40.15 & \multicolumn{1}{c|}{0.9837} & 41.21 & \multicolumn{1}{c|}{0.9864} & \textbf{41.25}         & \textbf{0.9864}         \\ \hline
		\multicolumn{2}{c|}{Average}                                                  & 22.53 & \multicolumn{1}{c|}{0.6555} & 28.58 & \multicolumn{1}{c|}{0.7651} & 30.89 & \multicolumn{1}{c|}{0.8135} & 31.71 & \multicolumn{1}{c|}{0.8470} & 32.39 & \multicolumn{1}{c|}{0.8554} & \textbf{32.47}         & \textbf{0.8567}         \\ \bottomrule
	\end{tabular}}
	\caption{Quantitative results on Set5 and Set14.}
	\label{Experiments1}
\end{table*}

\begin{table*}[t]\small
	\vspace{-1em}
	\centering
	\setlength{\tabcolsep}{6pt}{
	\begin{tabular}{cccccccccc}
		\toprule
		&                                 & \multicolumn{2}{c}{CSNet$^{+\S}$}          & \multicolumn{2}{c}{RK-CCSNet$^{\S}$}       & \multicolumn{2}{c}{MR-CCSNet}       & \multicolumn{2}{c}{MR-CCSNet$^{+}$} \\ \cline{3-10} 
		Data                                          & Ratio                           & PSNR  & SSIM                        & PSNR  & SSIM                        & PSNR  & SSIM                        & PSNR          & SSIM           \\ \hline
		\multicolumn{1}{c|}{\multirow{6}{*}{BSDS100}} & \multicolumn{1}{c|}{1.5625\%}  & 24.51 & \multicolumn{1}{c|}{0.6344} & 25.02 & \multicolumn{1}{c|}{0.6691} & 25.35 & \multicolumn{1}{c|}{0.6775} & \textbf{25.44}         & \textbf{0.6791}         \\
		\multicolumn{1}{c|}{}                         & \multicolumn{1}{c|}{3.125\%}  & 26.18 & \multicolumn{1}{c|}{0.7102} & 26.51 & \multicolumn{1}{c|}{0.7266} & 26.75 & \multicolumn{1}{c|}{0.7334} &\textbf{26.84}         & \textbf{0.7361}         \\
		\multicolumn{1}{c|}{}                         & \multicolumn{1}{c|}{6.25\%}  & 27.82 & \multicolumn{1}{c|}{0.7728} & 28.08 & \multicolumn{1}{c|}{0.7879} & 28.34 & \multicolumn{1}{c|}{0.7949} & \textbf{28.40}         & \textbf{0.7952}         \\
		\multicolumn{1}{c|}{}                         & \multicolumn{1}{c|}{12.5\%} & 29.77 & \multicolumn{1}{c|}{0.8424} & 29.98 & \multicolumn{1}{c|}{0.8559} & 30.39 & \multicolumn{1}{c|}{0.8632} & \textbf{30.43}         & \textbf{0.8639}         \\
		\multicolumn{1}{c|}{}                         & \multicolumn{1}{c|}{25\%} & 32.41 & \multicolumn{1}{c|}{0.9073} & 32.68 & \multicolumn{1}{c|}{0.9186} & 33.27 & \multicolumn{1}{c|}{0.9251} & \textbf{33.29}         & \textbf{0.9253}         \\
		\multicolumn{1}{c|}{}                         & \multicolumn{1}{c|}{50\%} & 36.21 & \multicolumn{1}{c|}{0.9582} & 37.29 & \multicolumn{1}{c|}{0.9695} & 38.03 & \multicolumn{1}{c|}{0.9731} & \textbf{38.07}         & \textbf{0.9732}        \\ \hline
		\multicolumn{2}{c|}{Average}                                                    & 29.48 & \multicolumn{1}{c|}{0.8042} & 29.93 & \multicolumn{1}{c|}{0.8213} & 30.36 & \multicolumn{1}{c|}{0.8279} & \textbf{30.41}         & \textbf{0.8288}         \\ \bottomrule
	\end{tabular}}
	\caption{Quantitative results on BSDS100.}
	\label{Experiments2}
\end{table*}

~\\
\noindent\textbf{Quantitative comparisons}\quad In \cref{Experiments1}, we report the quantitative comparisons on Set5 and Set14. The results show that MR-CCSNet and MR-CCSNet$^{+}$ are outperforms the four methods at all sampling ratios. Note that all DCS methods show a significant improvement comparing with the best traditional method, \ie GSR. Specifically, our model achieve the best performance in low sampling ratios. 
In average, MR-CCSNet$^{+}$ outperforms TVAL3, GSR, CSNet$^{+\S}$, and RK-CCSNet$^{\S}$ by \textbf{9.94dB}, \textbf{3.89dB}, \textbf{1.58dB}, and \textbf{0.76dB} in terms of PSNR, respectively, on Set5 and Set14. In addition, the average SSIM of MR-CCSNet$^{+}$ can be improved \textbf{0.2012}, \textbf{0.0916}, \textbf{0.0432}, and \textbf{0.0097}, respectively. We further compare MR-CCSNet and MR-CCSNet$^{+}$ with CSNet$^{+\S}$ and RK-CCSNet$^{\S}$ on BSDS100. \cref{Experiments2} show the evaluation results. It can be seen that both MR-CCSNet and MR-CCSNet$^{+}$ achieve a better reconstruction quality at all sampling ratios. Especially in the case of sampling ratio of 1.5625\%, MR-CCSNet$^{+}$ outperforms CSNet$^{+\S}$ and RK-CCSNet$^{\S}$ by \textbf{0.93dB} and \textbf{0.42dB}, respectively. Finally, we compare the performance of MR-CCSNet and MR-CCSNet$^{+}$. When the sampling ratio is 50\%, we observe that the PSNR and SSIM of them are very close. The reason is that GSM$^{+}$ degenerate into GSM in this case. As the sampling ratio decreases, MR-CCSNet$^{+}$ outperforms MR-CCSNet. The reason is that GSM$^{+}$ not only collects all level features, which is equivalent to GSM, but also extracts richer features for sampling and reconstructing. This is corresponding to our theoretical analysis in \cref{hscana}. All the experimental results demonstrate our model has state-of-the-art performance. \textbf{The more results are shown in \cref{appendix}}.

\begin{figure*}[t]
	\centering
	\includegraphics[width=0.95\linewidth]{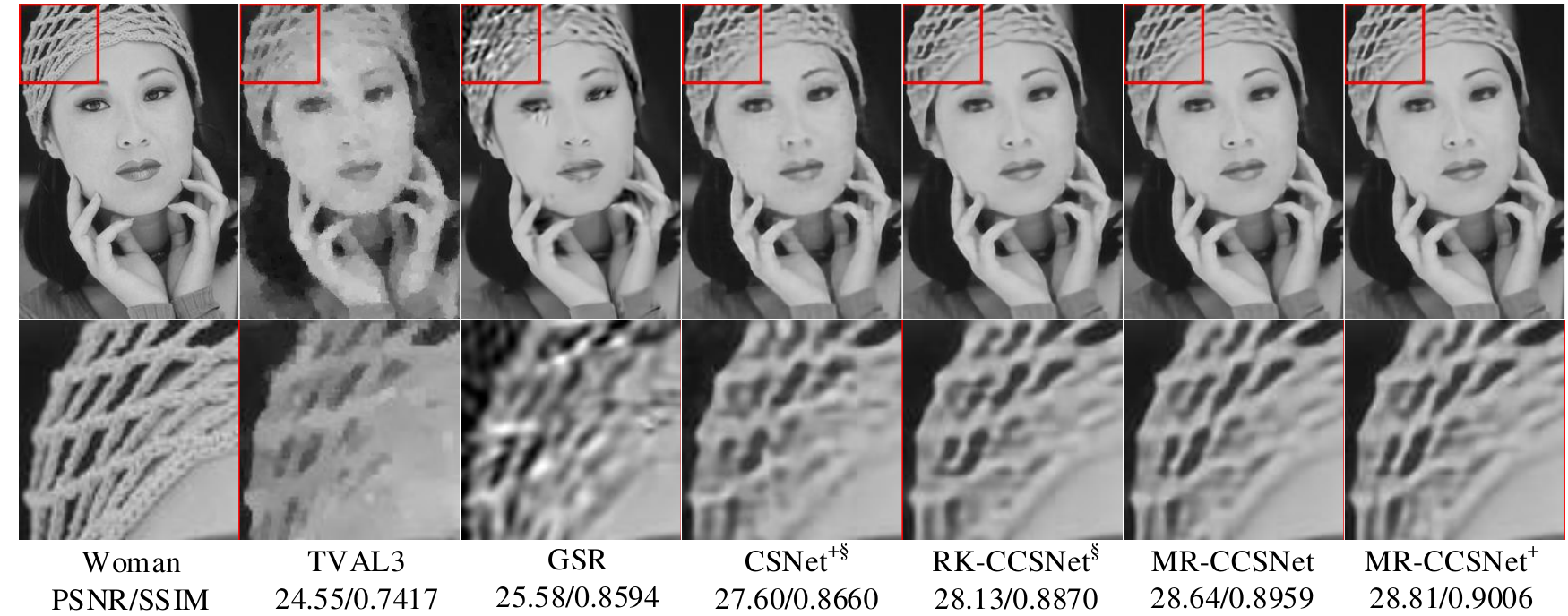}
	\caption{Visual comparisons of reconstructed image on Woman from Set5 in the sampling ratio of 6.25\%. }
	\label{exp1}
\end{figure*}

\begin{figure*}[t]
\vspace{-1em}
\centering
\includegraphics[width=0.95\linewidth]{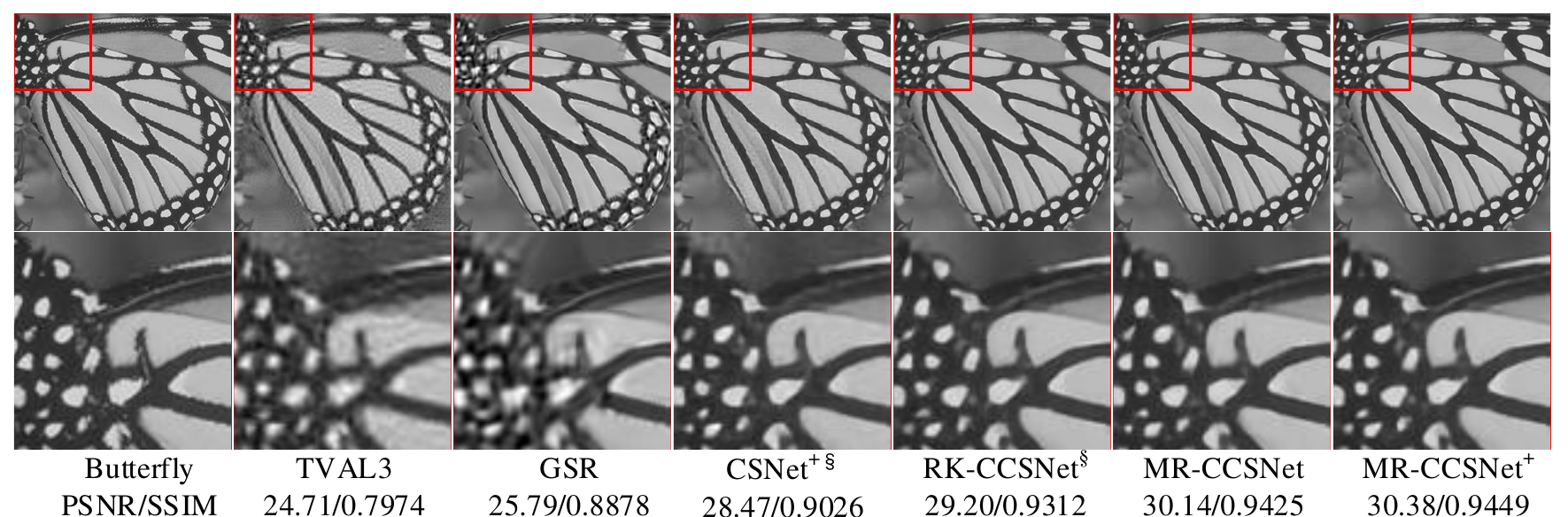}
\caption{Visual comparisons of reconstructed image on Butterfly from Set5 in the sampling ratio of 12.5\%. }
\label{exp2}
\end{figure*}

~\\
\noindent\textbf{Visual comparisons}\quad We also visually compare our method with the state-of-the-art image CS methods. We magnify the results in order to compare the reconstruction details. \cref{exp1} and \cref{exp2} show the visual comparisons in the case of sampling ratio of 6.25\% and 12.5\%, respectively. We can see that DCS methods can achieve higher reconstruction quality than traditional methods in extremely low sampling ratios. In addition, our model also recover finer details than DCS methods CSNet$^{+\S}$ and RK-CCSNet$^{+\S}$. For example, in the figure of Butterfly and Woman, it is obvious that our model is able to reconstruct texture details, which is smoother and sharper than other methods. This is mainly because the measurements in our model contain all level features where the low and mid-level features relate to the edges and complex textures in the image. In addition, extracting richer features by utilizing measurements multiple times also plays an important role.

\subsection{Running time comparison}	
In many practical applications, the running time is important and critical. \cref{time} shows the average running time on GPU/CPU for reconstructing a 256×256 image. The results of TVAL3 and GSR are taken from \cite{shi2019image}, and they are implemented with an Intel Core i7-3770 CPU. The running times of CSNet$^{+\S}$, RK-CCSNet$^{\S}$, MR-CCSNet, and MR-CCSNet$^{+}$ are implemented with an Intel Core i9-9900k CPU and a Nvidia RTX 2080Ti GPU. 
\begin{table}[h]\footnotesize
	\centering
	\setlength{\tabcolsep}{2mm}{
		\begin{tabular}{c|cc|cc}
			\toprule
			\multirow{2}{*}{Algorithm} & \multicolumn{2}{c|}{sampling ratio=0.01} & \multicolumn{2}{c}{sampling ratio=0.1} \\ \cline{2-5} 
			& CPU                  & GPU               & CPU                 & GPU              \\ \hline
			TVAL3                         & 2.3349               & -                 & 2.5871              & -                \\
			GSR                        & 235.6297             & -                 & 230.4755            & -                \\
			CSNet$^{+\S}$                      & -                    & 0.0075             & -                   & 0.0078           \\
			RK-CCSNet$^{\S}$                   & -                    & 0.0184            & -                   & 0.0181           \\
			MR-CCSNet                      & -                    & 0.0284            & -                   & 0.0272           \\
			MR-CCSNet$^{+}$                      & -                    & 0.0282            & -                   & 0.0271           \\
			 \bottomrule
	\end{tabular}}
	\caption{Average running time (in seconds).}
	\label{time}
\end{table}
It is obvious that traditional methods take about seconds to minutes to reconstruct the image. This is because they need multiple iterative operations during reconstruction. By comparison, the running time of DCS methods are improved by several orders of magnitude. The reason why our models slower than CSNet$^{+\S}$ and RK-CCSNet$^{\S}$ is MR-CCSNet$^{+}$ has more parameters. But it is more fast compared with traditional methods and achieves better reconstruction quality. We can see that the running time of MR-CCSNet and MR-CCSNet$^{+}$ are equal. This is because two models have approximately the same number of parameters.

\subsection{Ablation studies}
\begin{figure*}[t]
	\centering
	\includegraphics[width=.8\linewidth]{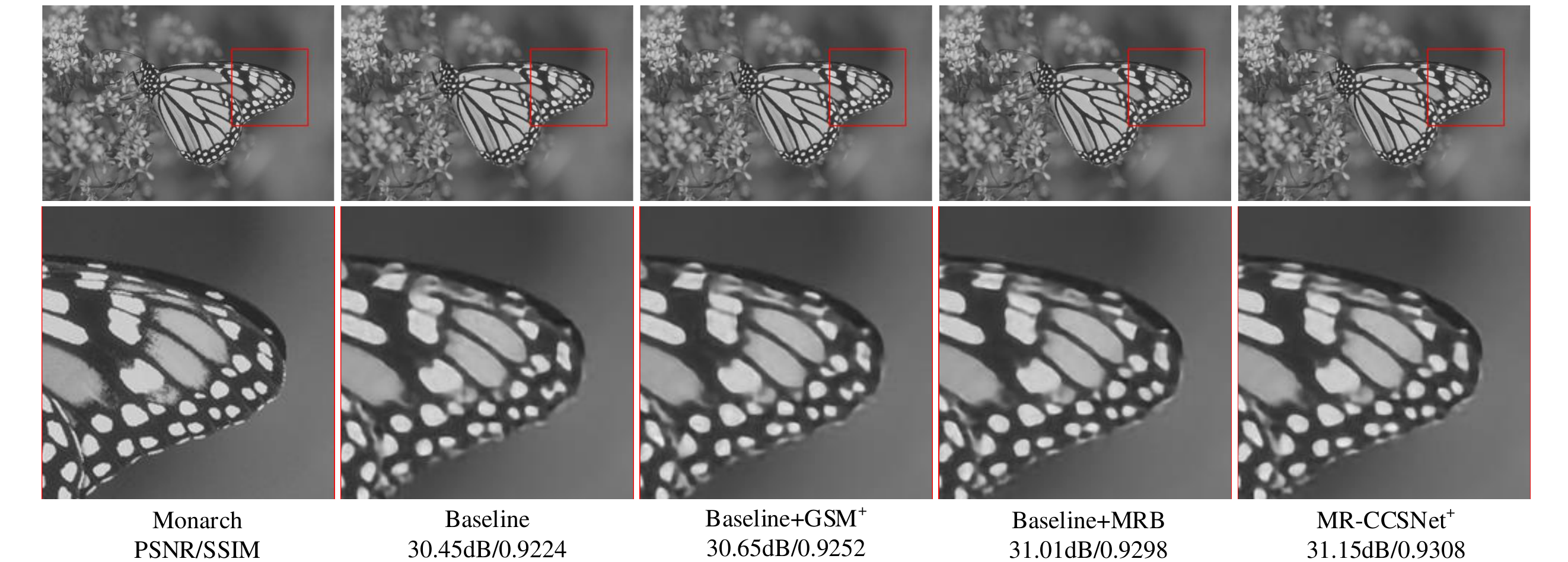}
	\caption{Visual comparisons of reconstructed image on Monarch from Set14 in the sampling ratio of 6.25\%. }
	\label{exp3}
\end{figure*}
\begin{table*}[t]\small
	\vspace{-1em}
	\centering
	\setlength{\tabcolsep}{1.5mm}{
		\begin{tabular}{cccccccccccccc}
			\toprule
			&                        & \multicolumn{2}{c}{1.5625\%}       & \multicolumn{2}{c}{3.125\%}       & \multicolumn{2}{c}{6.25\%}       & \multicolumn{2}{c}{12.5\%}      & \multicolumn{2}{c}{25\%}      & \multicolumn{2}{c}{50\%} \\ \cline{3-14} 
			GSM$^{+}$ & MRB                    & PSNR  & SSIM                        & PSNR  & SSIM                        & PSNR  & SSIM                        & PSNR  & SSIM                        & PSNR  & SSIM                        & PSNR          & SSIM           \\ \hline
			& \multicolumn{1}{c|}{}  & 25.02 & \multicolumn{1}{c|}{0.6691} & 26.51 & \multicolumn{1}{c|}{0.7266} & 28.08 & \multicolumn{1}{c|}{0.7879} & 29.98 & \multicolumn{1}{c|}{0.8559} & 32.68 & \multicolumn{1}{c|}{0.9186} & 37.29         & 0.9695         \\
			\checkmark   & \multicolumn{1}{c|}{}  & 25.14 & \multicolumn{1}{c|}{0.6737} & 26.58 & \multicolumn{1}{c|}{0.7281} & 28.18 & \multicolumn{1}{c|}{0.7915} & 30.15 & \multicolumn{1}{c|}{0.8591} & 32.88 & \multicolumn{1}{c|}{0.9211} & 37.71         & 0.9722         \\
			& \multicolumn{1}{c|}{\checkmark} & 25.29 & \multicolumn{1}{c|}{0.676}  & 26.61 & \multicolumn{1}{c|}{0.7307} & 28.26 & \multicolumn{1}{c|}{0.7931} & 30.27 & \multicolumn{1}{c|}{0.8614} & 33.01 & \multicolumn{1}{c|}{0.9216} & 37.53         & 0.9706         \\
			\checkmark   & \multicolumn{1}{c|}{\checkmark} & 25.49 & \multicolumn{1}{c|}{0.6811} & 26.88 & \multicolumn{1}{c|}{0.7359} & 28.38 & \multicolumn{1}{c|}{0.7955} & 30.36 & \multicolumn{1}{c|}{0.8629} & 33.24 & \multicolumn{1}{c|}{0.9248} & 37.98         & 0.9730         \\ \bottomrule
	\end{tabular}}
	\caption{The ablation studies of MR-CCSNet$^{+}$ on BSDS100.}
	\label{Ablation}
\end{table*}
In order to verify the efficacy of GSM$^{+}$ and MRB, we further conduct ablation studies on BSDS100. The models compared incude: Baseline (RK-CCSNet${^\S}$), Baseline with GSM$^{+}$, Baseline with MRB, and MR-CCSNet$^{+}$. From the results, as shown in \cref{Ablation}, we can observe that:

(1) Both GSM$^{+}$ and MRB are effective for improving the performance of reconstruction quality. This may be because GSM$^{+}$ can preserve more features in the image, and MRB can extract richer features for image reconstruction. 

(2) When the sampling ratio is low, MRB plays a more important role than GSM$^{+}$ for image reconstruction. Alternatively, GSM$^{+}$ plays a more important role than MRB when the sampling ratio is 50\%.

We also visually compare results of these four models, as shown in \cref{exp3}. The results is corresponding to our theoretical analysis. When the reconstruction algorithm is fixed, because GSM$^{+}$ takes advantage of the hierarchical nature of the network, the texture details of Baseline with GSM$^{+}$ smoother and sharper than Baseline. When the sensing network is fixed, because we utilize measurements in a deep manner, Baseline with MRB outperform Baseline. 
 
\section{Conclusion and future work}
In this paper, we propose Global Sensing Module and Measurements Reuse Block for image CS. GSM can take advantage of the hierarchical nature of the network for sampling. MRB can make full use of the measurements for improving the reconstructed image quality. In the experiments, we show that our model significantly and consistently outperforms state-of-the-art image CS methods. In particular, our methods also have good performances in extremely low sampling ratios. In addition, we demonstrate that GSM and MRB are effective by ablation studies. 

In the future, we will explore the following directions:

(1) In the sensing network, pooling operation loses information about the low-level features. We will explore a more effective way for collecting all level features.

(2) Attention mechanism can effectively help us in extracting matching features from measurements. We are interested in adding attention mechanism into MRB to improve its performance.

(3) In the real-world, because there are noise in the measurements, using them multiple times will introduce noise in the reconstruction process. We will explore how to improve the robustness for using measurements multiple times.

\noindent\textbf{Acknowledgement: }This research was supported by the National Natural Science Foundation of China (62173266).

{\small
\bibliographystyle{ieee_fullname}
\bibliography{egbib}

\begin{thebibliography}{10}\itemsep=-1pt

\bibitem{amini2011deterministic}
Arash Amini and Farokh Marvasti.
\newblock Deterministic construction of binary, bipolar, and ternary compressed
  sensing matrices.
\newblock {\em IEEE Transactions on Information Theory}, 57(4):2360--2370,
  2011.

\bibitem{arbelaez2010contour}
Pablo Arbelaez, Michael Maire, Charless Fowlkes, and Jitendra Malik.
\newblock Contour detection and hierarchical image segmentation.
\newblock {\em IEEE transactions on pattern analysis and machine intelligence},
  33(5):898--916, 2010.

\bibitem{bertero2020introduction}
Mario Bertero and Patrizia Boccacci.
\newblock {\em Introduction to inverse problems in imaging}.
\newblock CRC press, 2020.

\bibitem{bevilacqua2012low}
Marco Bevilacqua, Aline Roumy, Christine Guillemot, and Marie~Line
  Alberi-Morel.
\newblock Low-complexity single-image super-resolution based on nonnegative
  neighbor embedding.
\newblock 2012.

\bibitem{candes2008restricted}
Emmanuel~J Candes.
\newblock The restricted isometry property and its implications for compressed
  sensing.
\newblock {\em Comptes rendus mathematique}, 346(9-10):589--592, 2008.

\bibitem{candes2006near}
Emmanuel~J Candes and Terence Tao.
\newblock Near-optimal signal recovery from random projections: Universal
  encoding strategies?
\newblock {\em IEEE transactions on information theory}, 52(12):5406--5425,
  2006.

\bibitem{chen2011compressed}
Chen Chen, Eric~W Tramel, and James~E Fowler.
\newblock Compressed-sensing recovery of images and video using multihypothesis
  predictions.
\newblock In {\em 2011 conference record of the forty fifth asilomar conference
  on signals, systems and computers (ASILOMAR)}, pages 1193--1198. IEEE, 2011.

\bibitem{chen2001atomic}
Scott~Shaobing Chen, David~L Donoho, and Michael~A Saunders.
\newblock Atomic decomposition by basis pursuit.
\newblock {\em SIAM review}, 43(1):129--159, 2001.

\bibitem{daubechies2004iterative}
Ingrid Daubechies, Michel Defrise, and Christine De~Mol.
\newblock An iterative thresholding algorithm for linear inverse problems with
  a sparsity constraint.
\newblock {\em Communications on Pure and Applied Mathematics: A Journal Issued
  by the Courant Institute of Mathematical Sciences}, 57(11):1413--1457, 2004.

\bibitem{dinh2013measurement}
Khanh~Quoc Dinh, Hiuk~Jae Shim, and Byeungwoo Jeon.
\newblock Measurement coding for compressive imaging using a structural
  measuremnet matrix.
\newblock In {\em 2013 IEEE International Conference on Image Processing},
  pages 10--13. IEEE, 2013.

\bibitem{donoho2006compressed}
David~L Donoho.
\newblock Compressed sensing.
\newblock {\em IEEE Transactions on information theory}, 52(4):1289--1306,
  2006.

\bibitem{duarte2008single}
Marco~F Duarte, Mark~A Davenport, Dharmpal Takhar, Jason~N Laska, Ting Sun,
  Kevin~F Kelly, and Richard~G Baraniuk.
\newblock Single-pixel imaging via compressive sampling.
\newblock {\em IEEE signal processing magazine}, 25(2):83--91, 2008.

\bibitem{figueiredo2007gradient}
M{\'a}rio~AT Figueiredo, Robert~D Nowak, and Stephen~J Wright.
\newblock Gradient projection for sparse reconstruction: Application to
  compressed sensing and other inverse problems.
\newblock {\em IEEE Journal of selected topics in signal processing},
  1(4):586--597, 2007.

\bibitem{fowler2011multiscale}
James~E Fowler, Sungkwang Mun, and Eric~W Tramel.
\newblock Multiscale block compressed sensing with smoothed projected landweber
  reconstruction.
\newblock In {\em 2011 19th European Signal Processing Conference}, pages
  564--568. IEEE, 2011.

\bibitem{gan2007block}
Lu Gan.
\newblock Block compressed sensing of natural images.
\newblock In {\em 2007 15th International conference on digital signal
  processing}, pages 403--406. IEEE, 2007.

\bibitem{gao2015block}
Xinwei Gao, Jian Zhang, Wenbin Che, Xiaopeng Fan, and Debin Zhao.
\newblock Block-based compressive sensing coding of natural images by local
  structural measurement matrix.
\newblock In {\em 2015 Data Compression Conference}, pages 133--142. IEEE,
  2015.

\bibitem{haupt2006signal}
Jarvis Haupt and Robert Nowak.
\newblock Signal reconstruction from noisy random projections.
\newblock {\em IEEE Transactions on Information Theory}, 52(9):4036--4048,
  2006.

\bibitem{he2016deep}
Kaiming He, Xiangyu Zhang, Shaoqing Ren, and Jian Sun.
\newblock Deep residual learning for image recognition.
\newblock In {\em Proceedings of the IEEE conference on computer vision and
  pattern recognition}, pages 770--778, 2016.

\bibitem{hore2010image}
Alain Hore and Djemel Ziou.
\newblock Image quality metrics: Psnr vs. ssim.
\newblock In {\em 2010 20th international conference on pattern recognition},
  pages 2366--2369. IEEE, 2010.

\bibitem{kingma2014adam}
Diederik~P Kingma and Jimmy Ba.
\newblock Adam: A method for stochastic optimization.
\newblock {\em arXiv preprint arXiv:1412.6980}, 2014.

\bibitem{krizhevsky2012imagenet}
Alex Krizhevsky, Ilya Sutskever, and Geoffrey~E Hinton.
\newblock Imagenet classification with deep convolutional neural networks.
\newblock {\em Advances in neural information processing systems},
  25:1097--1105, 2012.

\bibitem{kulkarni2016reconnet}
Kuldeep Kulkarni, Suhas Lohit, Pavan Turaga, Ronan Kerviche, and Amit Ashok.
\newblock Reconnet: Non-iterative reconstruction of images from compressively
  sensed measurements.
\newblock In {\em Proceedings of the IEEE Conference on Computer Vision and
  Pattern Recognition}, pages 449--458, 2016.

\bibitem{li2013tval3}
C Li, W Yin, and Y Zhang.
\newblock Tval3: Tv minimization by augmented lagrangian and alternating
  direction agorithm 2009, 2013.

\bibitem{long2015fully}
Jonathan Long, Evan Shelhamer, and Trevor Darrell.
\newblock Fully convolutional networks for semantic segmentation.
\newblock In {\em Proceedings of the IEEE conference on computer vision and
  pattern recognition}, pages 3431--3440, 2015.

\bibitem{lu2017binary}
Weizhi Lu, Tao Dai, and Shu-Tao Xia.
\newblock Binary matrices for compressed sensing.
\newblock {\em IEEE Transactions on Signal Processing}, 66(1):77--85, 2017.

\bibitem{mallat1993matching}
St{\'e}phane~G Mallat and Zhifeng Zhang.
\newblock Matching pursuits with time-frequency dictionaries.
\newblock {\em IEEE Transactions on signal processing}, 41(12):3397--3415,
  1993.

\bibitem{mousavi2017learning}
Ali Mousavi and Richard~G Baraniuk.
\newblock Learning to invert: Signal recovery via deep convolutional networks.
\newblock In {\em 2017 IEEE international conference on acoustics, speech and
  signal processing (ICASSP)}, pages 2272--2276. IEEE, 2017.

\bibitem{mousavi2018data}
Ali Mousavi, Gautam Dasarathy, and Richard~G Baraniuk.
\newblock A data-driven and distributed approach to sparse signal
  representation and recovery.
\newblock In {\em International Conference on Learning Representations}, 2018.

\bibitem{mousavi2015deep}
Ali Mousavi, Ankit~B Patel, and Richard~G Baraniuk.
\newblock A deep learning approach to structured signal recovery.
\newblock In {\em 2015 53rd annual allerton conference on communication,
  control, and computing (Allerton)}, pages 1336--1343. IEEE, 2015.

\bibitem{mun2009block}
Sungkwang Mun and James~E Fowler.
\newblock Block compressed sensing of images using directional transforms.
\newblock In {\em 2009 16th IEEE international conference on image processing
  (ICIP)}, pages 3021--3024. IEEE, 2009.

\bibitem{mun2011residual}
Sungkwang Mun and James~E Fowler.
\newblock Residual reconstruction for block-based compressed sensing of video.
\newblock In {\em 2011 Data Compression Conference}, pages 183--192. IEEE,
  2011.

\bibitem{shi2019image}
Wuzhen Shi, Feng Jiang, Shaohui Liu, and Debin Zhao.
\newblock Image compressed sensing using convolutional neural network.
\newblock {\em IEEE Transactions on Image Processing}, 29:375--388, 2019.

\bibitem{shi2019scalable}
Wuzhen Shi, Feng Jiang, Shaohui Liu, and Debin Zhao.
\newblock Scalable convolutional neural network for image compressed sensing.
\newblock In {\em Proceedings of the IEEE/CVF Conference on Computer Vision and
  Pattern Recognition}, pages 12290--12299, 2019.

\bibitem{sun2020dual}
Yubao Sun, Jiwei Chen, Qingshan Liu, Bo Liu, and Guodong Guo.
\newblock Dual-path attention network for compressed sensing image
  reconstruction.
\newblock {\em IEEE Transactions on Image Processing}, 29:9482--9495, 2020.

\bibitem{tropp2007signal}
Joel~A Tropp and Anna~C Gilbert.
\newblock Signal recovery from random measurements via orthogonal matching
  pursuit.
\newblock {\em IEEE Transactions on information theory}, 53(12):4655--4666,
  2007.

\bibitem{wright2009sparse}
Stephen~J Wright, Robert~D Nowak, and M{\'a}rio~AT Figueiredo.
\newblock Sparse reconstruction by separable approximation.
\newblock {\em IEEE Transactions on signal processing}, 57(7):2479--2493, 2009.

\bibitem{xu2018lapran}
Kai Xu, Zhikang Zhang, and Fengbo Ren.
\newblock Lapran: A scalable laplacian pyramid reconstructive adversarial
  network for flexible compressive sensing reconstruction.
\newblock In {\em Proceedings of the European Conference on Computer Vision
  (ECCV)}, pages 485--500, 2018.

\bibitem{zeiler2014visualizing}
Matthew~D Zeiler and Rob Fergus.
\newblock Visualizing and understanding convolutional networks.
\newblock In {\em European conference on computer vision}, pages 818--833.
  Springer, 2014.

\bibitem{zeyde2010single}
Roman Zeyde, Michael Elad, and Matan Protter.
\newblock On single image scale-up using sparse-representations.
\newblock In {\em International conference on curves and surfaces}, pages
  711--730. Springer, 2010.

\bibitem{zhang2018ista}
Jian Zhang and Bernard Ghanem.
\newblock Ista-net: Interpretable optimization-inspired deep network for image
  compressive sensing.
\newblock In {\em Proceedings of the IEEE conference on computer vision and
  pattern recognition}, pages 1828--1837, 2018.

\bibitem{zhang2014group}
Jian Zhang, Debin Zhao, and Wen Gao.
\newblock Group-based sparse representation for image restoration.
\newblock {\em IEEE Transactions on Image Processing}, 23(8):3336--3351, 2014.

\bibitem{zhang2012image}
Jian Zhang, Debin Zhao, Chen Zhao, Ruiqin Xiong, Siwei Ma, and Wen Gao.
\newblock Image compressive sensing recovery via collaborative sparsity.
\newblock {\em IEEE Journal on Emerging and Selected Topics in Circuits and
  Systems}, 2(3):380--391, 2012.

\bibitem{zheng2020sequential}
Runkai Zheng, Yinqi Zhang, Daolang Huang, and Qingliang Chen.
\newblock Sequential convolution and runge-kutta residual architecture for
  image compressed sensing.
\newblock In {\em European Conference on Computer Vision}, pages 232--248.
  Springer, 2020.

\end{thebibliography}
}
\clearpage
\section{Appendix}
\label{appendix}
In \cref{all results set5 and set14} and \cref{all results bsds}, we show more results on Set5, Set14, and BSDS100. As previously mentioned, we use the code provided in \cite{zheng2020sequential} to train \textbf{CSNet$^{+\S}$} \cite{shi2019image} and \textbf{RK-CCSNet$^{\S}$}\cite{zheng2020sequential}. The results of \textbf{CSNet$^{+}$} and \textbf{RK-CCSNet} are taken from \cite{zheng2020sequential}.

\begin{table*}[t]\footnotesize
	\centering
	\setlength{\tabcolsep}{3pt}{
		\begin{tabular}{cccccccccccccccccc}
			\toprule
			&                                 & \multicolumn{2}{c}{TVAL3}           & \multicolumn{2}{c}{GSR}             & \multicolumn{2}{c}{CSNet$^{+\S}$}          & \multicolumn{2}{c}{CSNet$^{+}$}          & \multicolumn{2}{c}{RK-CCSNet$^{\S}$}       & \multicolumn{2}{c}{RK-CCSNet}       & \multicolumn{2}{c}{MR-CCSNet}       & \multicolumn{2}{c}{MR-CCSNet$^{+}$} \\ \cline{3-18} 
			Data                                        & Ratio                           & PSNR  & SSIM                        & PSNR  & SSIM                        & PSNR  & SSIM                        & PSNR  & SSIM                        & PSNR  & SSIM                        & PSNR  & SSIM                        & PSNR  & SSIM                        & PSNR          & SSIM           \\ \hline
			\multicolumn{1}{c|}{\multirow{6}{*}{Set5}}  & \multicolumn{1}{c|}{1.5625\%}  & 19.00 & \multicolumn{1}{c|}{0.4844} & 21.39 & \multicolumn{1}{c|}{0.5815} & 24.45 & \multicolumn{1}{c|}{0.6360} & 25.02 & \multicolumn{1}{c|}{0.6888} & 25.31 & \multicolumn{1}{c|}{0.7033} & 25.63 & \multicolumn{1}{c|}{0.7186} & 25.72 & \multicolumn{1}{c|}{0.7193} & 25.79         & 0.7189         \\
			\multicolumn{1}{c|}{}                       & \multicolumn{1}{c|}{3.125\%}  & 19.89 & \multicolumn{1}{c|}{0.5415} & 23.70 & \multicolumn{1}{c|}{0.6822} & 27.19 & \multicolumn{1}{c|}{0.7666} & 27.42 & \multicolumn{1}{c|}{0.7778} & 27.79 & \multicolumn{1}{c|}{0.8061} & 28.03 & \multicolumn{1}{c|}{0.8142} & 28.19 & \multicolumn{1}{c|}{0.8174} & 28.27         & 0.8208         \\
			\multicolumn{1}{c|}{}                       & \multicolumn{1}{c|}{6.25\%}  & 22.03 & \multicolumn{1}{c|}{0.6175} & 27.59 & \multicolumn{1}{c|}{0.8163} & 28.68 & \multicolumn{1}{c|}{0.8002} & 30.11 & \multicolumn{1}{c|}{0.8605} & 30.63 & \multicolumn{1}{c|}{0.8799} & 30.91 & \multicolumn{1}{c|}{0.8867} & 31.10 & \multicolumn{1}{c|}{0.8901} & 31.25         & 0.8918         \\
			\multicolumn{1}{c|}{}                       & \multicolumn{1}{c|}{12.5\%} & 23.75 & \multicolumn{1}{c|}{0.7365} & 31.61 & \multicolumn{1}{c|}{0.9016} & 33.55 & \multicolumn{1}{c|}{0.9243} & 33.57 & \multicolumn{1}{c|}{0.9250} & 34.27 & \multicolumn{1}{c|}{0.9393} & 35.05 & \multicolumn{1}{c|}{0.9461} & 35.03 & \multicolumn{1}{c|}{0.9464} & 35.16         & 0.9471         \\
			\multicolumn{1}{c|}{}                       & \multicolumn{1}{c|}{25\%} & 27.39 & \multicolumn{1}{c|}{0.8522} & 36.32 & \multicolumn{1}{c|}{0.9510} & 37.69 & \multicolumn{1}{c|}{0.9650} & 37.94 & \multicolumn{1}{c|}{0.9665} & 38.04 & \multicolumn{1}{c|}{0.9712} & 39.29 & \multicolumn{1}{c|}{0.9758} & 39.24 & \multicolumn{1}{c|}{0.9761} & 39.37         & 0.9766         \\
			\multicolumn{1}{c|}{}                       & \multicolumn{1}{c|}{50\%} & 33.11 & \multicolumn{1}{c|}{0.9430} & 42.18 & \multicolumn{1}{c|}{0.9908} & 42.54 & \multicolumn{1}{c|}{0.9852} & 42.70 & \multicolumn{1}{c|}{0.9856} & 43.90 & \multicolumn{1}{c|}{0.9901} & 44.72 & \multicolumn{1}{c|}{0.9913} & 45.07 & \multicolumn{1}{c|}{0.9919} & 45.11         & 0.9920         \\ \hline
			\multicolumn{1}{c|}{\multirow{6}{*}{Set14}} & \multicolumn{1}{c|}{1.5625\%}  & 16.79 & \multicolumn{1}{c|}{0.3993} & 18.93 & \multicolumn{1}{c|}{0.4399} & 22.78 & \multicolumn{1}{c|}{0.5369} & 23.13 & \multicolumn{1}{c|}{0.5768} & 23.36 & \multicolumn{1}{c|}{0.5917} & 23.32 & \multicolumn{1}{c|}{0.5933} & 23.61 & \multicolumn{1}{c|}{0.5993} & 23.69         & 0.6034         \\
			\multicolumn{1}{c|}{}                       & \multicolumn{1}{c|}{3.125\%}  & 18.40 & \multicolumn{1}{c|}{0.4514} & 20.26 & \multicolumn{1}{c|}{0.5184} & 24.96 & \multicolumn{1}{c|}{0.6602} & 25.03 & \multicolumn{1}{c|}{0.6660} & 25.26 & \multicolumn{1}{c|}{0.6914} & 25.42 & \multicolumn{1}{c|}{0.6968} & 25.56 & \multicolumn{1}{c|}{0.6997} & 25.63         & 0.7029         \\
			\multicolumn{1}{c|}{}                       & \multicolumn{1}{c|}{6.25\%}  & 19.65 & \multicolumn{1}{c|}{0.5287} & 23.59 & \multicolumn{1}{c|}{0.6526} & 26.33 & \multicolumn{1}{c|}{0.7178} & 27.25 & \multicolumn{1}{c|}{0.7651} & 27.24 & \multicolumn{1}{c|}{0.7836} & 27.48 & \multicolumn{1}{c|}{0.7897} & 27.91 & \multicolumn{1}{c|}{0.7986} & 28.00         & 0.7996         \\
			\multicolumn{1}{c|}{}                       & \multicolumn{1}{c|}{12.5\%} & 21.03 & \multicolumn{1}{c|}{0.6379} & 28.08 & \multicolumn{1}{c|}{0.7915} & 30.12 & \multicolumn{1}{c|}{0.8610} & 30.16 & \multicolumn{1}{c|}{0.8630} & 30.42 & \multicolumn{1}{c|}{0.8798} & 30.93 & \multicolumn{1}{c|}{0.8880} & 30.97 & \multicolumn{1}{c|}{0.8889} & 31.06         & 0.8898         \\
			\multicolumn{1}{c|}{}                       & \multicolumn{1}{c|}{25\%} & 22.69 & \multicolumn{1}{c|}{0.7731} & 31.82 & \multicolumn{1}{c|}{0.8939} & 33.81 & \multicolumn{1}{c|}{0.9339} & 33.92 & \multicolumn{1}{c|}{0.9354} & 34.16 & \multicolumn{1}{c|}{0.9443} & 35.03 & \multicolumn{1}{c|}{0.9505} & 35.04 & \multicolumn{1}{c|}{0.9510} & 35.11         & 0.9512         \\
			\multicolumn{1}{c|}{}                       & \multicolumn{1}{c|}{50\%} & 26.61 & \multicolumn{1}{c|}{0.9004} & 37.47 & \multicolumn{1}{c|}{0.9619} & 38.59 & \multicolumn{1}{c|}{0.9752} & 38.67 & \multicolumn{1}{c|}{0.9756} & 40.15 & \multicolumn{1}{c|}{0.9837} & 40.66 & \multicolumn{1}{c|}{0.9848} & 41.21 & \multicolumn{1}{c|}{0.9864} & 41.25         & 0.9864         \\ \hline
			\multicolumn{2}{c|}{Average}                                                  & 22.53 & \multicolumn{1}{c|}{0.6555} & 28.58 & \multicolumn{1}{c|}{0.7651} & 30.89 & \multicolumn{1}{c|}{0.8135} & 31.24 & 0.8322                      & 31.71 & \multicolumn{1}{c|}{0.8470} & 32.21 & 0.8530                      & 32.39 & \multicolumn{1}{c|}{0.8554} & 32.47         & 0.8567         \\ \bottomrule
	\end{tabular}}
	\caption{Quantitative results on Set5 and Set14.}
	\label{all results set5 and set14}
\end{table*}

\begin{table*}[t]\footnotesize
	\centering
	\setlength{\tabcolsep}{5pt}{
		\begin{tabular}{cccccccccccccc}
			\toprule
			&                                 & \multicolumn{2}{c}{CSNet$^{+\S}$}          & \multicolumn{2}{c}{CSNet$^{+}$}          & \multicolumn{2}{c}{RK-CCSNet$^{\S}$}       & \multicolumn{2}{c}{RK-CCSNet}       & \multicolumn{2}{c}{MR-CCSNet}       & \multicolumn{2}{c}{MR-CCSNet$^{+}$} \\ \cline{3-14} 
			Data                                          & Ratio                           & PSNR  & SSIM                        & PSNR  & SSIM                        & PSNR  & SSIM                        & PSNR  & SSIM                        & PSNR  & SSIM                        & PSNR          & SSIM           \\ \hline
			\multicolumn{1}{c|}{\multirow{6}{*}{BSDS100}} & \multicolumn{1}{c|}{1.5625\%}  & 24.51 & \multicolumn{1}{c|}{0.6344} & 25.01 & \multicolumn{1}{c|}{0.6904} & 25.02 & \multicolumn{1}{c|}{0.6691} & 25.56 & \multicolumn{1}{c|}{0.7055} & 25.35 & \multicolumn{1}{c|}{0.6775} & 25.44         & 0.6791         \\
			\multicolumn{1}{c|}{}                         & \multicolumn{1}{c|}{3.125\%}  & 26.18 & \multicolumn{1}{c|}{0.7102} & 26.55 & \multicolumn{1}{c|}{0.7413} & 26.51 & \multicolumn{1}{c|}{0.7266} & 26.99 & \multicolumn{1}{c|}{0.7564} & 26.75 & \multicolumn{1}{c|}{0.7334} & 26.84         & 0.7361         \\
			\multicolumn{1}{c|}{}                         & \multicolumn{1}{c|}{6.25\%}  & 27.82 & \multicolumn{1}{c|}{0.7728} & 28.14 & \multicolumn{1}{c|}{0.7977} & 28.08 & \multicolumn{1}{c|}{0.7879} & 28.60 & \multicolumn{1}{c|}{0.8133} & 28.34 & \multicolumn{1}{c|}{0.7949} & 28.40         & 0.7952         \\
			\multicolumn{1}{c|}{}                         & \multicolumn{1}{c|}{12.5\%} & 29.77 & \multicolumn{1}{c|}{0.8424} & 30.11 & \multicolumn{1}{c|}{0.8602} & 29.98 & \multicolumn{1}{c|}{0.8559} & 30.56 & \multicolumn{1}{c|}{0.8759} & 30.39 & \multicolumn{1}{c|}{0.8632} & 30.43         & 0.8639         \\
			\multicolumn{1}{c|}{}                         & \multicolumn{1}{c|}{25\%} & 32.41 & \multicolumn{1}{c|}{0.9073} & 32.81 & \multicolumn{1}{c|}{0.9206} & 32.68 & \multicolumn{1}{c|}{0.9186} & 33.43 & \multicolumn{1}{c|}{0.9335} & 33.27 & \multicolumn{1}{c|}{0.9251} & 33.29         & 0.9253         \\
			\multicolumn{1}{c|}{}                         & \multicolumn{1}{c|}{50\%} & 36.21 & \multicolumn{1}{c|}{0.9582} & 36.62 & \multicolumn{1}{c|}{0.9659} & 37.29 & \multicolumn{1}{c|}{0.9695} & 37.92 & \multicolumn{1}{c|}{0.9766} & 38.03 & \multicolumn{1}{c|}{0.9731} & 38.07         & 0.9732         \\ \hline
			\multicolumn{2}{c}{Average}                                                     & 29.48 & 0.8042                      & 29.87 & 0.8294                      & 29.93 & 0.8213                      & 30.51 & 0.8435                      & 30.36 & 0.8279                      & 30.41         & 0.8288         \\ \bottomrule
	\end{tabular}}
	\caption{Quantitative results on BSDS100.}
	\label{all results bsds}
\end{table*}

\end{document}